\documentclass[11pt,a4paper]{article}
\usepackage[hyperref]{acl2019}
\usepackage{times}
\usepackage{latexsym}
\usepackage{graphicx}
\usepackage{url}
\usepackage{balance}
\usepackage{longtable}
\usepackage{booktabs}

\aclfinalcopy 

\title{Using Word Embeddings to Examine Gender Bias in Dutch Newspapers, 1950-1990}

\author{Melvin Wevers \\
  DHLab KNAW Humanities Cluster\\
  Oudezijds Achterburgwal 185 \\
  1012DK Amsterdam, the Netherlands \\
  \texttt{melvin.wevers@dh.huc.knaw.nl}}

\date{}

\begin{document}
\maketitle
\begin{abstract}
Contemporary debates on filter bubbles and polarization in public and social media raise the question to what extent news media of the past exhibited biases. This paper specifically examines bias related to gender in six Dutch national newspapers between 1950 and 1990. We measure bias related to gender by comparing local changes in word embedding models trained on newspapers with divergent ideological backgrounds.
We demonstrate clear differences in gender bias and changes within and between newspapers over time. In relation to themes such as sexuality and leisure, we see the bias moving toward women, whereas, generally, the bias shifts in the direction of men, despite growing female employment number and feminist movements. Even though Dutch society became less stratified ideologically (depillarization), we found an increasing divergence in gender bias between religious and social-democratic on the one hand and liberal newspapers on the other.
Methodologically, this paper illustrates how word embeddings can be used to examine historical language change. Future work will investigate how fine-tuning deep contextualized embedding models, such as ELMO, might be used for similar tasks with greater contextual information.
\end{abstract}

\section{Introduction}
In recent years, public and academic debates about the possible impact of filter bubbles and the role of polarization in public and social media have been widespread~\cite{pariser2011filter, flaxman2016filter}. In these debates, news media have been described as belonging to particular political ideologies, producing skewed views on topics, such as climate change or immigration. These contemporary debates raise the question to what extent newspapers in the past operated in filter bubbles driven by their own ideological bias.

This paper examines gender bias in historical newspapers. By looking at differences in the strength of association between male and female dimensions of gender on the one hand, and words that represent occupations, psychological states, or social life, on the other, we examine the gender bias in and between several Dutch newspapers over time. Did certain newspapers exhibit a bias toward men or women in relationship to specific aspects of society, behavior, or culture?

Newspapers are an excellent source to study societal debates. They function as a transceiver; both the producer and the messenger of public discourse ~\cite{schudson_power_1982}. Margaret Marshall~\shortcite{marshall_contesting_1995} claims that researchers can uncover the ``values, assumptions, and concerns, and ways of thinking that were a part of the public discourse of that time'' by analyzing ``the arguments, language, the discourse practices that inhabit the pages of public magazines, newspapers, and early professional journals.'' 

The period 1950-1990 is of particular interest as Dutch society underwent clear industrialization and modernization as well as ideological shifts ~\cite{schot_technology_2010}. After the Second World War, Dutch society was stratified according to ideological and religious ``pillars'', a phenomenon known as pillarization. These pillars can be categorized as Catholic, Protestant, socialist, and liberal ~\cite{wintle_economic_2000}. 
Newspapers were often aligned to one of these pillars ~\cite{wijfjes_journalistiek_2004, rooij_kranten:_1974}. The newspaper \textit{Trouw}, for example, has a distinct Protestant origin, while \textit{Volkskrant} and \textit{De Telegraaf} can be characterized as, respectively, Catholic and neutral. In recent years, the latter transformed into a newspaper with clear conservative leanings. Newspaper historians have studied the ideological backgrounds of Dutch newspapers using traditional hermeneutic means to which this study adds a computational analysis of language use related to gender. 

The representation of gender in public discourse is related to ideological struggles over gender equality. Several feminist waves materialized in the Netherlands. The origins of the first feminist wave can be traced back to the mid-nineteenth century and lasted until the interwar period. It took until the 1960s for feminism to flare up again in the Netherlands. In between, confessional parties were vocal in their anti-feminist policies. During the 1960s, the second feminist wave, also known as `new feminism', focused on gender equality in areas such as work, education, sexuality, marriage, and family \cite{ribberink_feminisme_1987}. 

\begin{figure}
  \includegraphics[width=\linewidth]{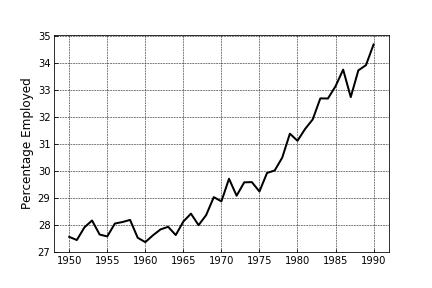}
  \caption{Female Employment Numbers}
  \label{fig:female_employment}
\end{figure}

The increasing equality between men and women is reflected in growing female employment numbers, which increased from 27.5 percent in 1950 to almost 35 percent in 1990 (Figure~\ref{fig:female_employment}).\footnote{\url{https://opendata.cbs.nl/statline/\#/CBS/nl/}} Apart from Scandinavia, the Netherlands has the highest levels of equality in Europe. Nonetheless, in terms of education and employment, women are still lagging behind and reports of gender discrimination are not uncommon in the Netherlands ~\cite{baali_overheid_2018, ministerie_van_onderwijs_vrouwenemancipatie_2009}.

\section{Related Work}
Word embedding models can be used for a wide range of lexical-semantic tasks \cite{baroni2014don, kulkarni_statistically_2015}. Hamilton et al.~\shortcite{hamilton_cultural_2016-1} show how word embeddings can also be used to measure semantic shifts by comparing the contexts in which words are used to denote continuity and changes in language use. More recent work focused on the role of bias in word embeddings, specifically bias related to politics, gender, and ethnicity \cite{azarbonyad_words_2017, bolukbasi_quantifying_2016, garg_word_2018}. Gonen et al.~\shortcite{gonen_lipstick_2019} demonstrate that debiasing methods work, but argue that we should not remove them. Azarbonyad et al.~\shortcite{azarbonyad_words_2017} compare semantic spaces related to political views in the UK parliament, effectively comparing biases between embeddings. Garg et al.~\shortcite{garg_word_2018} turn to biases in embedding to study shifts related to gender and ethnicity.
 
This study builds upon the work of Garg et al. \shortcite{garg_word_2018}, and applies it to the context of the Netherlands---represented by Dutch newspapers. We extend their method further by distinguishing between sources, rather than using a comprehensive gold standard data set. We also incorporate external lexicons, such as the emotion lexicon from Cornetto, the \textit{Nederlandse Voornamenbank} (database of Dutch first names), the Dutch translation of LIWC (Linguistic Inquiry and Word Count) and HISCO (Historical International Classification of Occupations)~\cite{vossen_cornetto_2007, tausczik2010psychological, boot_dutch_2017, zijdeman_hsn_2013, bloothooft_nederlandse_2010}.

\section{Data}
The data set consists of six Dutch national newspapers: \textit{NRC Handelsblad (NRC)}, \textit{Het Vrije Volk (VV)}, \textit{Parool}, \textit{Telegraaf}, \textit{Trouw}, and \textit{Volkskrant (VK)}.\footnote{ The digitized newspapers were provided by the National Library of the Netherlands. http://www.delpher.nl} These newspapers can be characterized ideologically as liberal, social-democratic, liberal, neutral/conservative, Protestant, and Catholic.

For the analysis, we rely on the articles and not the advertisements in the newspapers. We preprocess the text by removing stopwords, punctuation, numerical characters, and words shorter than three and longer than fifteen characters. 
The quality of the digitized text varies throughout the corpus due to imperfections in the original material and limitations of the recognition software. Because of the variations in OCR quality, we only retain words that also appeared in a Dutch dictionary.

\begin{figure}
  \includegraphics[width=\linewidth]{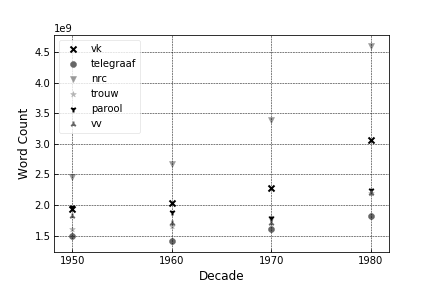}
  \caption{Total number of words per embedding model}
  \label{fig:vocab-size}
\end{figure}

We use the Gensim implementation of Word2Vec to train four embedding models per newspaper, each representing one decade between 1950 and 1990.\footnote{\url{https://radimrehurek.com/gensim/}} The models were trained using C-BOW with hierarchical softmax, with a dimensionality of 300, a minimal word count and context of 5, and downsampling of $10^{-5}$.\footnote{Code can be found here: \url{https://github.com/melvinwevers/historical_concepts} and the models here: \url{http://doi.org/10.5281/zenodo.3237380}} Figure \ref{fig:vocab-size} shows that the size of the vocabulary approximately doubles for some newspapers between 1950 and 1990. The variance of the targets words, however, was small ($\mu \approx 0.003$) and constant ($\sigma [1.3^{-9}, 2.9^{-9}$]), indicating model stability. Since we calculate bias relative to each model, these differences in vocabulary size will have little impact on shifts in bias.

To measure gender bias, we use three sets of targets words. First, we extract a list of approximately 12.5k job titles from the HISCO data set. Second, we select emotion words with a confidence score of 1.0, a positive polarity above 0.5 ($n = 476$) and a negative polarity below -0.5 ($n = 636$) from Cornetto. Third, we rely on the Dutch translation of LIWC2001, which contains lists of words to measure psychological and cognitive states~\cite{pennebaker2001linguistic}. We use the following LIWC (sub)categories: Affective and Emotional Processes; Cognitive Processes; Sensory and Perceptual Processes; Social Processes; Occupation; Leisure activity; Money and Financial Issues; Metaphysical Issues; and Physical states.

\section{Methodology}
\begin{figure}
  \includegraphics[width=\linewidth]{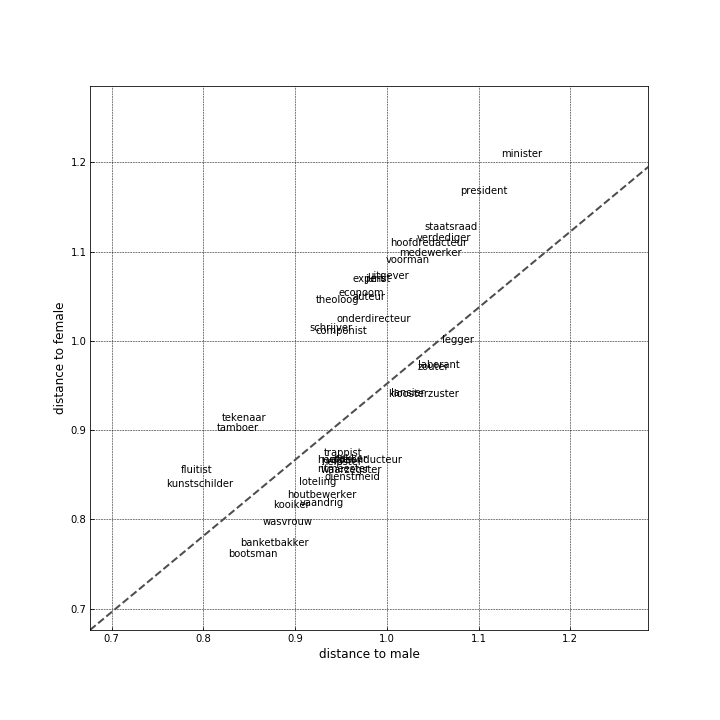}
  \vspace*{-10mm}
  \caption{Job titles with strong bias towards men and women in \textit{De Volkskrant}, 1980-1990}
  \label{fig:example}
\end{figure}

For the calculation of gender bias, we construct two vectors representing the gender dimensions (male, female). We do this by creating an average vector that includes words referring to male (`man', `his', `father', etc.) or female as well as the most popular first names in the Netherlands for the period 1950-1990.\footnote{The word lists for both vectors can be found in Appendix A. The first names were harvested from \url{https://www.meertens.knaw.nl/nvb/}}
Next, we calculate the distance between each gender vector and every word in a list of target words, for example, words that denote occupations: a greater distance indicates that a word is less closely associated with that dimension of gender. The difference between the distances for both gender vectors represents the gender bias: positive meaning a bias toward women and negative toward men. Figure \ref{fig:example} shows the biases related to forty job titles. Words above the diagonal are biased towards men, and those underneath the diagonal towards women. 

Finally, after standardizing and centering the bias values, we apply Bayesian linear regression to determine whether the bias changed over time. The linear model is formulated as:
\[\mu_i = \alpha + \beta * Y_{i} + \epsilon,\]
with $\mu_{i}$ the bias for each decade ($i$) and $Y_{i}$ the coefficient related to each decade ($i$). The likelihood function is: $X \sim \mathcal{N}(\mu, \sigma)$ with priors defined: $\alpha \sim \mathcal{N}(0, 2)$, $\beta \sim \mathcal{N}(0, 2)$, and $\epsilon \sim \mathrm{HalfCauchy}(\beta = 1)$. For model training, we use a No-U-Turn-Sampler (NUTS) (5k draws, 1.5k tuning steps, Highest Posterior Density (HPD) of .95).\footnote{HPD is the Bayesian equivalent of the frequentists confidence interval in Frequentist credible interval. \url{https://docs.pymc.io}} For the target words Job Titles, the proposed model (Model B) outperforms a model that only includes the intercept (Model A), indicating that bias changes as a function of time (Table \ref{table:model1} \& Table \ref{table:model2}). 

\begin{table}
\tiny
\begin{tabular}{llllllll}
\toprule
{} &     WAIC & pWAIC &  dWAIC & weight &      SE &   dSE \\
\midrule
Model B &  64624.8 &   2.9 &      0 &   0.99 &   201.6 &     0 \\
Model A &  64682.1 &  1.88 &  57.28 &   0.01 &  201.36 &  15.2 \\
\bottomrule
\end{tabular}
\caption{Model Comparison}
\label{table:model1}	
\end{table}

\begin{table}
\tiny
\begin{tabular}{lrrrrrrr}
\toprule
{} &   mean &     sd &  hpd\_2.5 &  hpd\_97.5 &     n\_eff &   Rhat \\
\midrule
a     & -0.164 &  0.010 &       -0.185 &    -0.145 &  1315.073 &  1.000 \\
bY    &  0.046 &  0.006 &       0.033 &     0.055 &  1261.437 &  0.999 \\
sigma &  1.001 &  0.005 &       0.992 &     1.010 &  1035.282 &  1.003 \\
\bottomrule
\end{tabular}
\caption{Model B Summary}
\label{table:model2}
\end{table}

We compute a linear model that combines all newspapers for the target words Job Titles, Positive Emotions, Negative Emotions, and the selected LIWC columns. Then, for the same categories, we compute individual linear models for each newspaper. The resulting models are reported in Appendix B.
\section{Results}
The combined linear models, including all newspapers, generally display minimal shifts in bias. While the effects are weak, they fall within a .95 HPD. Partly, the weak trends are related to opposing shifts in the individual newspapers, cancelling each other out. Nonetheless, the bias associated with the categories `TV', `Music', `Metaphysical issues', `Sexuality' navigate toward women (0.22, 0.12, 0.15, 0.22), with all of them starting from a position that was clearly oriented toward men (-0.36, -0.20, -0.28, -0.39).\footnote{Numbers refer to the slope} 
Conversely, `Money', `Grooming', and Negative Emotion words move toward men (-0.24, -0.17, -0.16), which in the 1950s were all more closely related to women (0.33, 0.20, 0.19). For the Job Titles, we see a slight move toward women (0.05), while words from the LIWC category Occupation move marginally in the direction of men (-0.05). This suggests that job titles might be more closely related to women, while the notion of working gravitates toward men. 
\begin{figure}
  \includegraphics[width=\linewidth]{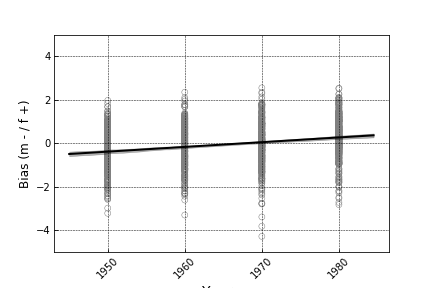}
  \caption{Combined model `Sexuality'}
  \label{fig:overall_sexuality}
\end{figure}

The linear models for the individual newspapers demonstrate distinct differences between the newspapers. First, \textit{Volkskrant} is the most stable newspapers with 56\% of the categories not changing.\footnote{Lower confidence interval $<$ 0 and upper $>$ 0} When bias changes in this newspaper, it moves toward women 9 out the 11 categories that change. \textit{Telegraaf}, \textit{NRC}, and \textit{Parool} generally move toward men, respectively (84\%, 92\%, and 80\%). The bias of \textit{Trouw} and \textit{Vrije Volk}, contrarily, move toward women (both 72\%). 

A noteworthy result is that in all newspapers the bias shifts toward men in the category `money'. Moreover, they also all exhibit a move toward women for the category `sexuality', with the clearest shift in \textit{Volkskrant}, \textit{Trouw}, and \textit{Vrije Volk}. 
\begin{figure}
  \includegraphics[width=\linewidth]{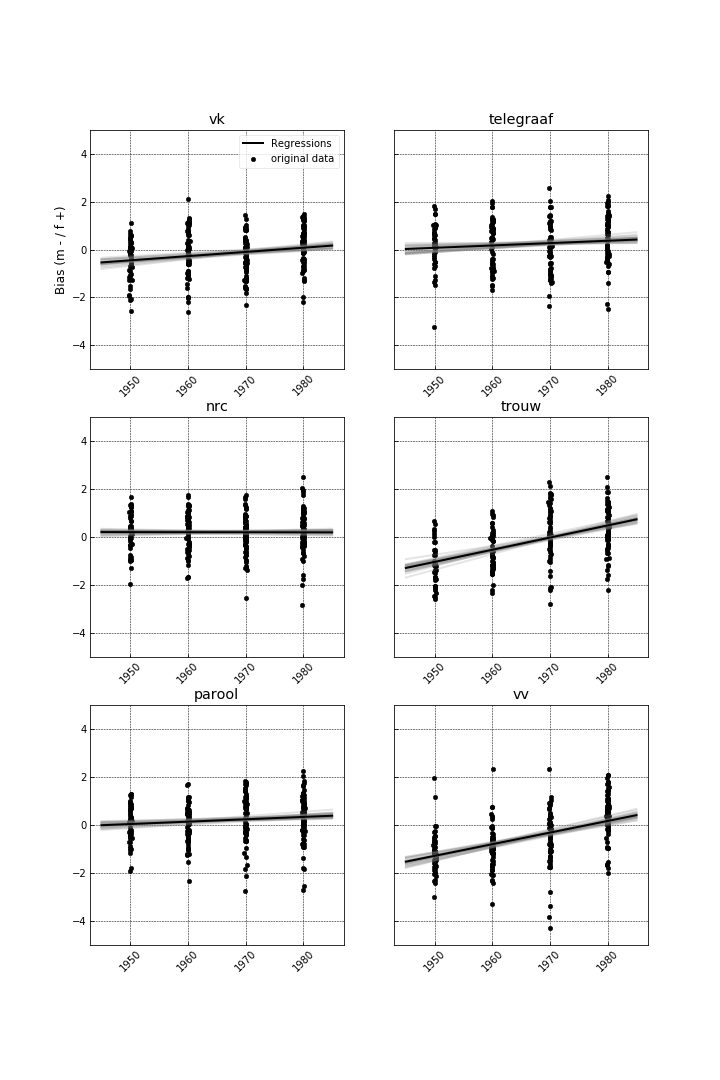}
  \vspace*{-15mm}
  \caption{Individual newspaper model `Sexuality'}
  \label{fig:indiv_sexuality}
\end{figure}

\section{Discussion}
While the newspaper discourse as a whole is fairly stable, individual newspapers show clear divergences with regard to their bias and changes in this bias. We see that the newspapers with a social-democratic (\textit{Vrije Volk}) and religious background, either Catholic (\textit{Volkskrant}) and Protestant (\textit{Trouw}) demonstrate the clearest shift in bias toward women. The liberal/conservative newspapers \textit{Telegraaf}, \textit{NRC Handelsblad}, and \textit{Parool}, on the contrary, orient themselves more clearly toward men. 
Despite increasing female employment numbers in the Netherlands, the association with job titles moves only gradually toward women, while words associated with working move toward men. 
More detailed analysis of the individual trend within each decade is necessary to untangle what exactly is taking place. For example, which words show the biggest shift, and can we identify groups of associated words of which particular words show divergent behavior?
Methodologically, this paper shows how word embedding models can be used to trace general shifts in language related to gender. Nevertheless, certain cultural expressions of gender are not captured by distributional semantics represented through word embeddings, but rather in syntax, for example, through the use of active of passive sentences. Future work will investigate how fine-tuning state-of-the-art embedding models, such as ELMO and BERT, can be leveraged to gain more contextual knowledge about words and their association with gender~\cite{peters_deep_2018}.  

\section*{Acknowledgments}
I would like to thank Folgert Karsdorp for his feedback. This research was part the project ``Digital Humanities Approaches to Reference Cultures: The Emergence of the United States in Public Discourse in the Netherlands, 1890-1990'', which was funded by the Dutch Research Council (NWO).\\

\bibliography{acl2019}
\bibliographystyle{acl_natbib}

\clearpage

\appendix
\section{Gender Vectors}
\textbf{Male vector}: hij (he), vader (father), opa (grandpa), zoon (son), man (man), mannen (men), \& heer (sir)

\textbf{Female vector}: zij (she), moeder (mother), oma (grandma), dochter (daughter), vrouw (woman), vrouwen (women), \& dame (madam)

\clearpage
\onecolumn
\section{Linear Models}
\tiny
\begin{longtable}{llrrrr}
\toprule
      &    &   mean &     sd &  hpd\_2.5 &  hpd\_97.5 \\
\midrule
positive\_words & a & -0.009 &  0.018 &   -0.042 &     0.026 \\
      & bY & -0.023 &  0.010 &   -0.042 &    -0.004 \\
negative\_words & a &  0.193 &  0.017 &    0.158 &     0.224 \\
      & bY & -0.157 &  0.009 &   -0.175 &    -0.141 \\
job\_titles & a & -0.164 &  0.011 &   -0.185 &    -0.142 \\
      & bY &  0.046 &  0.006 &    0.035 &     0.059 \\
Affect & a &  0.198 &  0.023 &    0.156 &     0.241 \\
      & bY & -0.162 &  0.012 &   -0.185 &    -0.139 \\
Posemo & a &  0.104 &  0.022 &    0.064 &     0.147 \\
      & bY & -0.098 &  0.012 &   -0.121 &    -0.076 \\
Negemo & a &  0.251 &  0.024 &    0.203 &     0.296 \\
      & bY & -0.194 &  0.012 &   -0.218 &    -0.171 \\
Anx & a &  0.309 &  0.027 &    0.256 &     0.357 \\
      & bY & -0.232 &  0.015 &   -0.261 &    -0.203 \\
Anger & a &  0.184 &  0.027 &    0.130 &     0.236 \\
      & bY & -0.150 &  0.014 &   -0.174 &    -0.121 \\
Sad & a &  0.209 &  0.026 &    0.156 &     0.254 \\
      & bY & -0.171 &  0.013 &   -0.198 &    -0.147 \\
Senses & a &  0.134 &  0.023 &    0.090 &     0.183 \\
      & bY & -0.112 &  0.012 &   -0.137 &    -0.089 \\
Social & a &  0.033 &  0.023 &   -0.011 &     0.080 \\
      & bY & -0.042 &  0.012 &   -0.066 &    -0.018 \\
Occup & a &  0.035 &  0.022 &   -0.010 &     0.076 \\
      & bY & -0.053 &  0.012 &   -0.074 &    -0.030 \\
Leisure & a & -0.066 &  0.025 &   -0.114 &    -0.022 \\
      & bY &  0.031 &  0.013 &    0.007 &     0.055 \\
Home & a & -0.027 &  0.043 &   -0.105 &     0.062 \\
      & bY & -0.001 &  0.023 &   -0.046 &     0.043 \\
Sports & a &  0.045 &  0.038 &   -0.038 &     0.105 \\
      & bY & -0.042 &  0.020 &   -0.080 &    -0.002 \\
TV & a & -0.364 &  0.088 &   -0.526 &    -0.195 \\
      & bY &  0.217 &  0.045 &    0.130 &     0.302 \\
Music & a & -0.200 &  0.049 &   -0.292 &    -0.102 \\
      & bY &  0.122 &  0.025 &    0.076 &     0.168 \\
Money & a &  0.335 &  0.028 &    0.284 &     0.390 \\
      & bY & -0.243 &  0.015 &   -0.272 &    -0.215 \\
Metaph & a & -0.281 &  0.030 &   -0.341 &    -0.225 \\
      & bY &  0.146 &  0.015 &    0.119 &     0.180 \\
Physcal & a & -0.008 &  0.027 &   -0.063 &     0.041 \\
      & bY & -0.020 &  0.014 &   -0.044 &     0.007 \\
Body & a &  0.043 &  0.025 &   -0.010 &     0.087 \\
      & bY & -0.059 &  0.013 &   -0.084 &    -0.034 \\
Sexual & a & -0.382 &  0.046 &   -0.471 &    -0.289 \\
      & bY &  0.216 &  0.023 &    0.167 &     0.255 \\
Eating & a & -0.007 &  0.034 &   -0.069 &     0.055 \\
      & bY & -0.015 &  0.018 &   -0.046 &     0.023 \\
Sleep & a &  0.134 &  0.049 &    0.041 &     0.230 \\
      & bY & -0.110 &  0.027 &   -0.160 &    -0.054 \\
Groom & a &  0.204 &  0.055 &    0.088 &     0.300 \\
      & bY & -0.166 &  0.031 &   -0.224 &    -0.105 \\
\bottomrule
\caption{Combined Linear Model}
\end{longtable}
\clearpage
\twocolumn

\tiny
\clearpage
\onecolumn
\begin{longtable}{llrrrrl}

\toprule
   &    &   mean &     sd &  hpd\_2.5 &  hpd\_97.5 &        category \\
\midrule
nrc & a &  0.649 &  0.049 &    0.567 &     0.758 &          Affect \\
   & a &  0.572 &  0.049 &    0.482 &     0.667 &          Posemo \\
   & a &  0.701 &  0.050 &    0.605 &     0.800 &          Negemo \\
   & a &  0.797 &  0.054 &    0.684 &     0.901 &             Anx \\
   & a &  0.687 &  0.050 &    0.592 &     0.787 &           Anger \\
   & a &  0.648 &  0.055 &    0.553 &     0.761 &             Sad \\
   & a &  0.631 &  0.044 &    0.545 &     0.711 &          Senses \\
   & a &  0.474 &  0.050 &    0.379 &     0.567 &          Social \\
   & a &  0.480 &  0.045 &    0.386 &     0.561 &           Occup \\
   & a &  0.485 &  0.047 &    0.401 &     0.577 &         Leisure \\
   & a &  0.465 &  0.095 &    0.288 &     0.653 &            Home \\
   & a &  0.487 &  0.075 &    0.325 &     0.621 &          Sports \\
   & a &  0.290 &  0.158 &   -0.018 &     0.585 &              TV \\
   & a &  0.645 &  0.093 &    0.478 &     0.829 &           Music \\
   & a &  0.719 &  0.051 &    0.622 &     0.810 &           Money \\
   & a &  0.159 &  0.060 &    0.049 &     0.278 &          Metaph \\
   & a &  0.559 &  0.055 &    0.441 &     0.657 &         Physcal \\
   & a &  0.571 &  0.051 &    0.476 &     0.666 &            Body \\
   & a &  0.184 &  0.094 &   -0.015 &     0.343 &          Sexual \\
   & a &  0.661 &  0.067 &    0.526 &     0.784 &          Eating \\
   & a &  0.799 &  0.088 &    0.639 &     0.966 &           Sleep \\
   & a &  0.461 &  0.116 &    0.184 &     0.653 &           Groom \\
   & a &  0.451 &  0.035 &    0.376 &     0.515 &  positive\_words \\
   & a &  0.662 &  0.032 &    0.604 &     0.720 &  negative\_words \\
   & a &  0.181 &  0.022 &    0.134 &     0.222 &      job\_titles \\
   & bY & -0.384 &  0.026 &   -0.431 &    -0.331 &          Affect \\
   & bY & -0.308 &  0.026 &   -0.359 &    -0.258 &          Posemo \\
   & bY & -0.413 &  0.027 &   -0.462 &    -0.359 &          Negemo \\
   & bY & -0.486 &  0.029 &   -0.543 &    -0.431 &             Anx \\
   & bY & -0.395 &  0.027 &   -0.446 &    -0.343 &           Anger \\
   & bY & -0.369 &  0.027 &   -0.420 &    -0.315 &             Sad \\
   & bY & -0.356 &  0.023 &   -0.397 &    -0.306 &          Senses \\
   & bY & -0.275 &  0.025 &   -0.327 &    -0.226 &          Social \\
   & bY & -0.271 &  0.024 &   -0.316 &    -0.220 &           Occup \\
   & bY & -0.207 &  0.026 &   -0.255 &    -0.159 &         Leisure \\
   & bY & -0.345 &  0.051 &   -0.439 &    -0.250 &            Home \\
   & bY & -0.171 &  0.037 &   -0.247 &    -0.101 &          Sports \\
   & bY & -0.090 &  0.084 &   -0.240 &     0.074 &              TV \\
   & bY & -0.219 &  0.050 &   -0.304 &    -0.115 &           Music \\
   & bY & -0.487 &  0.026 &   -0.536 &    -0.436 &           Money \\
   & bY & -0.125 &  0.032 &   -0.189 &    -0.066 &          Metaph \\
   & bY & -0.303 &  0.027 &   -0.351 &    -0.248 &         Physcal \\
   & bY & -0.308 &  0.027 &   -0.360 &    -0.261 &            Body \\
   & bY &  0.012 &  0.046 &   -0.073 &     0.098 &          Sexual \\
   & bY & -0.409 &  0.037 &   -0.476 &    -0.336 &          Eating \\
   & bY & -0.461 &  0.047 &   -0.558 &    -0.375 &           Sleep \\
   & bY & -0.283 &  0.063 &   -0.413 &    -0.162 &           Groom \\
   & bY & -0.244 &  0.018 &   -0.277 &    -0.204 &  positive\_words \\
   & bY & -0.384 &  0.017 &   -0.415 &    -0.348 &  negative\_words \\
   & bY & -0.131 &  0.012 &   -0.154 &    -0.106 &      job\_titles \\
parool & a &  0.602 &  0.051 &    0.515 &     0.701 &          Affect \\
   & a &  0.594 &  0.049 &    0.496 &     0.686 &          Posemo \\
   & a &  0.648 &  0.053 &    0.552 &     0.755 &          Negemo \\
   & a &  0.763 &  0.060 &    0.648 &     0.883 &             Anx \\
   & a &  0.604 &  0.055 &    0.487 &     0.706 &           Anger \\
   & a &  0.627 &  0.059 &    0.525 &     0.749 &             Sad \\
   & a &  0.544 &  0.052 &    0.438 &     0.648 &          Senses \\
   & a &  0.322 &  0.056 &    0.214 &     0.424 &          Social \\
   & a &  0.434 &  0.059 &    0.316 &     0.545 &           Occup \\
   & a &  0.296 &  0.057 &    0.191 &     0.413 &         Leisure \\
   & a &  0.196 &  0.106 &   -0.016 &     0.374 &            Home \\
   & a &  0.387 &  0.093 &    0.207 &     0.556 &          Sports \\
   & a &  0.211 &  0.188 &   -0.113 &     0.584 &              TV \\
   & a &  0.310 &  0.106 &    0.114 &     0.529 &           Music \\
   & a &  0.759 &  0.060 &    0.641 &     0.878 &           Money \\
   & a &  0.026 &  0.067 &   -0.106 &     0.154 &          Metaph \\
   & a &  0.349 &  0.058 &    0.249 &     0.473 &         Physcal \\
   & a &  0.400 &  0.056 &    0.303 &     0.512 &            Body \\
   & a &  0.049 &  0.103 &   -0.124 &     0.259 &          Sexual \\
   & a &  0.361 &  0.074 &    0.206 &     0.485 &          Eating \\
   & a &  0.515 &  0.104 &    0.320 &     0.714 &           Sleep \\
   & a &  0.418 &  0.136 &    0.184 &     0.679 &           Groom \\
   & a &  0.479 &  0.041 &    0.401 &     0.554 &  positive\_words \\
   & a &  0.638 &  0.038 &    0.563 &     0.716 &  negative\_words \\
   & a & -0.053 &  0.025 &   -0.097 &    -0.001 &      job\_titles \\
   & bY & -0.341 &  0.027 &   -0.390 &    -0.290 &          Affect \\
   & bY & -0.335 &  0.027 &   -0.387 &    -0.283 &          Posemo \\
   & bY & -0.377 &  0.028 &   -0.429 &    -0.324 &          Negemo \\
   & bY & -0.463 &  0.033 &   -0.524 &    -0.398 &             Anx \\
   & bY & -0.341 &  0.030 &   -0.400 &    -0.284 &           Anger \\
   & bY & -0.388 &  0.031 &   -0.451 &    -0.330 &             Sad \\
   & bY & -0.310 &  0.028 &   -0.367 &    -0.260 &          Senses \\
   & bY & -0.159 &  0.030 &   -0.216 &    -0.107 &          Social \\
   & bY & -0.254 &  0.032 &   -0.319 &    -0.195 &           Occup \\
   & bY & -0.145 &  0.030 &   -0.200 &    -0.083 &         Leisure \\
   & bY & -0.141 &  0.056 &   -0.256 &    -0.035 &            Home \\
   & bY & -0.245 &  0.048 &   -0.332 &    -0.150 &          Sports \\
   & bY &  0.001 &  0.097 &   -0.172 &     0.208 &              TV \\
   & bY & -0.063 &  0.057 &   -0.160 &     0.056 &           Music \\
   & bY & -0.490 &  0.031 &   -0.549 &    -0.424 &           Money \\
   & bY &  0.109 &  0.034 &    0.047 &     0.177 &          Metaph \\
   & bY & -0.159 &  0.031 &   -0.214 &    -0.100 &         Physcal \\
   & bY & -0.214 &  0.029 &   -0.278 &    -0.161 &            Body \\
   & bY &  0.091 &  0.054 &   -0.004 &     0.194 &          Sexual \\
   & bY & -0.213 &  0.039 &   -0.285 &    -0.139 &          Eating \\
   & bY & -0.276 &  0.058 &   -0.380 &    -0.154 &           Sleep \\
   & bY & -0.283 &  0.074 &   -0.417 &    -0.136 &           Groom \\
   & bY & -0.273 &  0.022 &   -0.313 &    -0.229 &  positive\_words \\
   & bY & -0.378 &  0.019 &   -0.417 &    -0.346 &  negative\_words \\
   & bY & -0.009 &  0.014 &   -0.032 &     0.022 &      job\_titles \\
telegraaf & a &  0.606 &  0.062 &    0.496 &     0.729 &          Affect \\
   & a &  0.502 &  0.061 &    0.388 &     0.622 &          Posemo \\
   & a &  0.649 &  0.067 &    0.527 &     0.784 &          Negemo \\
   & a &  0.847 &  0.076 &    0.704 &     1.013 &             Anx \\
   & a &  0.479 &  0.066 &    0.344 &     0.594 &           Anger \\
   & a &  0.575 &  0.068 &    0.452 &     0.709 &             Sad \\
   & a &  0.609 &  0.057 &    0.492 &     0.710 &          Senses \\
   & a &  0.487 &  0.063 &    0.363 &     0.606 &          Social \\
   & a &  0.086 &  0.063 &   -0.037 &     0.205 &           Occup \\
   & a &  0.178 &  0.075 &    0.021 &     0.307 &         Leisure \\
   & a &  0.685 &  0.127 &    0.412 &     0.933 &            Home \\
   & a &  0.112 &  0.112 &   -0.107 &     0.341 &          Sports \\
   & a &  0.047 &  0.182 &   -0.310 &     0.404 &              TV \\
   & a & -0.062 &  0.127 &   -0.289 &     0.188 &           Music \\
   & a &  0.487 &  0.075 &    0.342 &     0.642 &           Money \\
   & a &  0.289 &  0.072 &    0.150 &     0.428 &          Metaph \\
   & a &  0.398 &  0.067 &    0.261 &     0.526 &         Physcal \\
   & a &  0.369 &  0.066 &    0.227 &     0.482 &            Body \\
   & a &  0.116 &  0.122 &   -0.121 &     0.355 &          Sexual \\
   & a &  0.547 &  0.089 &    0.367 &     0.712 &          Eating \\
   & a &  0.877 &  0.112 &    0.669 &     1.097 &           Sleep \\
   & a &  0.584 &  0.144 &    0.295 &     0.855 &           Groom \\
   & a &  0.335 &  0.048 &    0.252 &     0.435 &  positive\_words \\
   & a &  0.631 &  0.046 &    0.542 &     0.717 &  negative\_words \\
   & a & -0.020 &  0.030 &   -0.079 &     0.039 &      job\_titles \\
   & bY & -0.298 &  0.032 &   -0.358 &    -0.231 &          Affect \\
   & bY & -0.190 &  0.033 &   -0.248 &    -0.124 &          Posemo \\
   & bY & -0.337 &  0.037 &   -0.409 &    -0.265 &          Negemo \\
   & bY & -0.384 &  0.040 &   -0.462 &    -0.313 &             Anx \\
   & bY & -0.278 &  0.036 &   -0.344 &    -0.207 &           Anger \\
   & bY & -0.260 &  0.035 &   -0.329 &    -0.194 &             Sad \\
   & bY & -0.272 &  0.028 &   -0.329 &    -0.219 &          Senses \\
   & bY & -0.147 &  0.034 &   -0.207 &    -0.079 &          Social \\
   & bY & -0.120 &  0.033 &   -0.193 &    -0.065 &           Occup \\
   & bY & -0.080 &  0.038 &   -0.149 &    -0.002 &         Leisure \\
   & bY & -0.195 &  0.067 &   -0.333 &    -0.058 &            Home \\
   & bY & -0.176 &  0.059 &   -0.291 &    -0.059 &          Sports \\
   & bY &  0.131 &  0.096 &   -0.049 &     0.340 &              TV \\
   & bY &  0.085 &  0.066 &   -0.031 &     0.205 &           Music \\
   & bY & -0.254 &  0.039 &   -0.344 &    -0.181 &           Money \\
   & bY &  0.004 &  0.039 &   -0.072 &     0.081 &          Metaph \\
   & bY & -0.165 &  0.035 &   -0.233 &    -0.098 &         Physcal \\
   & bY & -0.223 &  0.036 &   -0.287 &    -0.150 &            Body \\
   & bY &  0.080 &  0.065 &   -0.039 &     0.203 &          Sexual \\
   & bY & -0.200 &  0.045 &   -0.280 &    -0.103 &          Eating \\
   & bY & -0.275 &  0.061 &   -0.373 &    -0.140 &           Sleep \\
   & bY & -0.365 &  0.079 &   -0.532 &    -0.224 &           Groom \\
   & bY & -0.140 &  0.026 &   -0.194 &    -0.093 &  positive\_words \\
   & bY & -0.317 &  0.024 &   -0.362 &    -0.273 &  negative\_words \\
   & bY & -0.049 &  0.016 &   -0.077 &    -0.017 &      job\_titles \\
trouw & a & -0.089 &  0.059 &   -0.192 &     0.032 &          Affect \\
   & a & -0.234 &  0.048 &   -0.331 &    -0.149 &          Posemo \\
   & a & -0.025 &  0.061 &   -0.138 &     0.103 &          Negemo \\
   & a &  0.009 &  0.062 &   -0.102 &     0.125 &             Anx \\
   & a & -0.158 &  0.066 &   -0.270 &    -0.024 &           Anger \\
   & a & -0.038 &  0.064 &   -0.152 &     0.086 &             Sad \\
   & a & -0.295 &  0.055 &   -0.400 &    -0.189 &          Senses \\
   & a & -0.273 &  0.052 &   -0.366 &    -0.163 &          Social \\
   & a & -0.068 &  0.054 &   -0.172 &     0.040 &           Occup \\
   & a & -0.273 &  0.058 &   -0.379 &    -0.170 &         Leisure \\
   & a & -0.429 &  0.125 &   -0.665 &    -0.183 &            Home \\
   & a &  0.131 &  0.101 &   -0.069 &     0.316 &          Sports \\
   & a & -0.865 &  0.159 &   -1.184 &    -0.549 &              TV \\
   & a & -0.640 &  0.107 &   -0.853 &    -0.441 &           Music \\
   & a &  0.092 &  0.057 &   -0.018 &     0.203 &           Money \\
   & a & -0.795 &  0.064 &   -0.915 &    -0.671 &          Metaph \\
   & a & -0.406 &  0.061 &   -0.519 &    -0.292 &         Physcal \\
   & a & -0.250 &  0.070 &   -0.386 &    -0.110 &            Body \\
   & a & -1.038 &  0.126 &   -1.272 &    -0.808 &          Sexual \\
   & a & -0.576 &  0.090 &   -0.756 &    -0.411 &          Eating \\
   & a & -0.319 &  0.112 &   -0.532 &    -0.091 &           Sleep \\
   & a & -0.103 &  0.153 &   -0.438 &     0.172 &           Groom \\
   & a & -0.279 &  0.043 &   -0.361 &    -0.192 &  positive\_words \\
   & a & -0.150 &  0.042 &   -0.246 &    -0.084 &  negative\_words \\
   & a & -0.404 &  0.028 &   -0.460 &    -0.354 &      job\_titles \\
   & bY &  0.051 &  0.030 &   -0.010 &     0.107 &          Affect \\
   & bY &  0.113 &  0.026 &    0.067 &     0.163 &          Posemo \\
   & bY &  0.036 &  0.033 &   -0.029 &     0.097 &          Negemo \\
   & bY &  0.047 &  0.033 &   -0.017 &     0.109 &             Anx \\
   & bY &  0.115 &  0.034 &    0.056 &     0.184 &           Anger \\
   & bY &  0.021 &  0.034 &   -0.043 &     0.087 &             Sad \\
   & bY &  0.191 &  0.028 &    0.144 &     0.250 &          Senses \\
   & bY &  0.142 &  0.027 &    0.089 &     0.198 &          Social \\
   & bY &  0.074 &  0.030 &    0.017 &     0.131 &           Occup \\
   & bY &  0.229 &  0.030 &    0.171 &     0.281 &         Leisure \\
   & bY &  0.326 &  0.066 &    0.203 &     0.453 &            Home \\
   & bY &  0.060 &  0.054 &   -0.039 &     0.171 &          Sports \\
   & bY &  0.544 &  0.078 &    0.386 &     0.689 &              TV \\
   & bY &  0.350 &  0.055 &    0.240 &     0.463 &           Music \\
   & bY &  0.006 &  0.031 &   -0.050 &     0.062 &           Money \\
   & bY &  0.343 &  0.034 &    0.283 &     0.410 &          Metaph \\
   & bY &  0.301 &  0.033 &    0.235 &     0.366 &         Physcal \\
   & bY &  0.221 &  0.036 &    0.152 &     0.297 &            Body \\
   & bY &  0.503 &  0.061 &    0.375 &     0.604 &          Sexual \\
   & bY &  0.434 &  0.046 &    0.346 &     0.519 &          Eating \\
   & bY &  0.252 &  0.059 &    0.132 &     0.363 &           Sleep \\
   & bY &  0.125 &  0.080 &   -0.048 &     0.273 &           Groom \\
   & bY &  0.206 &  0.022 &    0.164 &     0.250 &  positive\_words \\
   & bY &  0.115 &  0.021 &    0.074 &     0.153 &  negative\_words \\
   & bY &  0.207 &  0.015 &    0.180 &     0.236 &      job\_titles \\
vk & a & -0.136 &  0.049 &   -0.224 &    -0.040 &          Affect \\
   & a & -0.211 &  0.047 &   -0.312 &    -0.128 &          Posemo \\
   & a & -0.098 &  0.055 &   -0.202 &     0.014 &          Negemo \\
   & a & -0.070 &  0.063 &   -0.183 &     0.053 &             Anx \\
   & a & -0.033 &  0.052 &   -0.130 &     0.071 &           Anger \\
   & a & -0.080 &  0.051 &   -0.170 &     0.027 &             Sad \\
   & a & -0.075 &  0.049 &   -0.168 &     0.021 &          Senses \\
   & a & -0.190 &  0.055 &   -0.293 &    -0.086 &          Social \\
   & a & -0.176 &  0.054 &   -0.277 &    -0.062 &           Occup \\
   & a & -0.100 &  0.057 &   -0.206 &     0.015 &         Leisure \\
   & a & -0.137 &  0.105 &   -0.326 &     0.086 &            Home \\
   & a & -0.053 &  0.097 &   -0.246 &     0.128 &          Sports \\
   & a & -0.539 &  0.179 &   -0.852 &    -0.197 &              TV \\
   & a & -0.063 &  0.109 &   -0.272 &     0.146 &           Music \\
   & a &  0.103 &  0.059 &   -0.011 &     0.219 &           Money \\
   & a & -0.483 &  0.064 &   -0.612 &    -0.364 &          Metaph \\
   & a & -0.100 &  0.056 &   -0.194 &     0.031 &         Physcal \\
   & a & -0.066 &  0.058 &   -0.166 &     0.052 &            Body \\
   & a & -0.454 &  0.099 &   -0.630 &    -0.266 &          Sexual \\
   & a & -0.118 &  0.074 &   -0.255 &     0.018 &          Eating \\
   & a & -0.120 &  0.095 &   -0.301 &     0.064 &           Sleep \\
   & a &  0.371 &  0.132 &    0.127 &     0.617 &           Groom \\
   & a & -0.218 &  0.038 &   -0.284 &    -0.139 &  positive\_words \\
   & a & -0.066 &  0.035 &   -0.134 &    -0.003 &  negative\_words \\
   & a & -0.030 &  0.026 &   -0.082 &     0.020 &      job\_titles \\
   & bY &  0.016 &  0.025 &   -0.029 &     0.064 &          Affect \\
   & bY &  0.067 &  0.027 &    0.011 &     0.115 &          Posemo \\
   & bY & -0.015 &  0.029 &   -0.066 &     0.045 &          Negemo \\
   & bY & -0.058 &  0.033 &   -0.121 &     0.011 &             Anx \\
   & bY & -0.028 &  0.028 &   -0.075 &     0.032 &           Anger \\
   & bY &  0.011 &  0.026 &   -0.036 &     0.070 &             Sad \\
   & bY &  0.019 &  0.027 &   -0.032 &     0.071 &          Senses \\
   & bY &  0.047 &  0.029 &   -0.006 &     0.105 &          Social \\
   & bY &  0.114 &  0.028 &    0.058 &     0.169 &           Occup \\
   & bY &  0.110 &  0.031 &    0.049 &     0.168 &         Leisure \\
   & bY &  0.090 &  0.056 &   -0.014 &     0.197 &            Home \\
   & bY &  0.124 &  0.051 &    0.027 &     0.223 &          Sports \\
   & bY &  0.256 &  0.093 &    0.077 &     0.421 &              TV \\
   & bY &  0.101 &  0.056 &   -0.003 &     0.219 &           Music \\
   & bY & -0.087 &  0.032 &   -0.153 &    -0.027 &           Money \\
   & bY &  0.190 &  0.033 &    0.123 &     0.249 &          Metaph \\
   & bY &  0.026 &  0.029 &   -0.029 &     0.083 &         Physcal \\
   & bY &  0.039 &  0.029 &   -0.015 &     0.096 &            Body \\
   & bY &  0.177 &  0.049 &    0.070 &     0.256 &          Sexual \\
   & bY &  0.023 &  0.039 &   -0.046 &     0.102 &          Eating \\
   & bY & -0.004 &  0.053 &   -0.107 &     0.102 &           Sleep \\
   & bY & -0.196 &  0.070 &   -0.326 &    -0.060 &           Groom \\
   & bY &  0.118 &  0.020 &    0.081 &     0.157 &  positive\_words \\
   & bY & -0.018 &  0.018 &   -0.049 &     0.019 &  negative\_words \\
   & bY &  0.028 &  0.013 &    0.003 &     0.053 &      job\_titles \\
vv & a & -0.480 &  0.057 &   -0.589 &    -0.370 &          Affect \\
   & a & -0.640 &  0.055 &   -0.752 &    -0.531 &          Posemo \\
   & a & -0.381 &  0.064 &   -0.503 &    -0.261 &          Negemo \\
   & a & -0.479 &  0.065 &   -0.600 &    -0.345 &             Anx \\
   & a & -0.503 &  0.061 &   -0.618 &    -0.382 &           Anger \\
   & a & -0.500 &  0.063 &   -0.615 &    -0.372 &             Sad \\
   & a & -0.616 &  0.052 &   -0.724 &    -0.521 &          Senses \\
   & a & -0.633 &  0.056 &   -0.750 &    -0.527 &          Social \\
   & a & -0.575 &  0.059 &   -0.699 &    -0.478 &           Occup \\
   & a & -0.987 &  0.068 &   -1.107 &    -0.850 &         Leisure \\
   & a & -0.939 &  0.108 &   -1.145 &    -0.722 &            Home \\
   & a & -0.756 &  0.102 &   -0.942 &    -0.555 &          Sports \\
   & a & -1.403 &  0.226 &   -1.836 &    -0.950 &              TV \\
   & a & -1.427 &  0.102 &   -1.625 &    -1.234 &           Music \\
   & a & -0.172 &  0.065 &   -0.294 &    -0.053 &           Money \\
   & a & -0.919 &  0.068 &   -1.046 &    -0.781 &          Metaph \\
   & a & -0.880 &  0.066 &   -1.003 &    -0.752 &         Physcal \\
   & a & -0.779 &  0.067 &   -0.917 &    -0.668 &            Body \\
   & a & -1.326 &  0.111 &   -1.532 &    -1.107 &          Sexual \\
   & a & -0.921 &  0.077 &   -1.048 &    -0.730 &          Eating \\
   & a & -0.992 &  0.107 &   -1.184 &    -0.764 &           Sleep \\
   & a & -0.549 &  0.151 &   -0.828 &    -0.255 &           Groom \\
   & a & -0.867 &  0.044 &   -0.950 &    -0.781 &  positive\_words \\
   & a & -0.633 &  0.039 &   -0.704 &    -0.558 &  negative\_words \\
   & a & -0.667 &  0.023 &   -0.707 &    -0.617 &      job\_titles \\
   & bY &  0.006 &  0.031 &   -0.051 &     0.065 &          Affect \\
   & bY &  0.085 &  0.029 &    0.026 &     0.139 &          Posemo \\
   & bY & -0.049 &  0.034 &   -0.113 &     0.010 &          Negemo \\
   & bY & -0.050 &  0.034 &   -0.118 &     0.016 &             Anx \\
   & bY &  0.030 &  0.032 &   -0.034 &     0.090 &           Anger \\
   & bY & -0.027 &  0.034 &   -0.093 &     0.041 &             Sad \\
   & bY &  0.063 &  0.028 &    0.007 &     0.116 &          Senses \\
   & bY &  0.152 &  0.030 &    0.090 &     0.208 &          Social \\
   & bY &  0.146 &  0.030 &    0.090 &     0.205 &           Occup \\
   & bY &  0.280 &  0.035 &    0.209 &     0.342 &         Leisure \\
   & bY &  0.266 &  0.059 &    0.154 &     0.372 &            Home \\
   & bY &  0.136 &  0.054 &    0.030 &     0.241 &          Sports \\
   & bY &  0.475 &  0.114 &    0.274 &     0.724 &              TV \\
   & bY &  0.494 &  0.055 &    0.394 &     0.604 &           Music \\
   & bY & -0.123 &  0.035 &   -0.188 &    -0.053 &           Money \\
   & bY &  0.388 &  0.035 &    0.319 &     0.459 &          Metaph \\
   & bY &  0.204 &  0.034 &    0.133 &     0.274 &         Physcal \\
   & bY &  0.145 &  0.036 &    0.083 &     0.220 &            Body \\
   & bY &  0.507 &  0.056 &    0.410 &     0.615 &          Sexual \\
   & bY &  0.295 &  0.041 &    0.221 &     0.384 &          Eating \\
   & bY &  0.143 &  0.057 &    0.039 &     0.267 &           Sleep \\
   & bY &  0.023 &  0.083 &   -0.127 &     0.203 &           Groom \\
   & bY &  0.217 &  0.023 &    0.179 &     0.266 &  positive\_words \\
   & bY &  0.069 &  0.021 &    0.029 &     0.109 &  negative\_words \\
   & bY &  0.242 &  0.013 &    0.218 &     0.269 &      job\_titles \\
\bottomrule
\caption{Individual Linear Model}
\end{longtable}
\clearpage
\twocolumn

\end{document}